\documentclass[mathematics,article,submit,moreauthors]{Definitions/mdpi} 
\def\today{September 20, 2022}
\renewcommand{\linenumbers}{}

\firstpage{1} 
\pubvolume{1}
\issuenum{1}
\articlenumber{0}
\pubyear{2022}
\copyrightyear{2022}
\datereceived{} 
\dateaccepted{} 
\datepublished{} 
\hreflink{https://doi.org/} 

\usepackage{bold-extra}
\usepackage{stmaryrd}

\usepackage{pdfpages}
\usepackage{morefloats}
\usepackage{graphics}
\usepackage{bbm}
\usepackage{epsfig}
\usepackage{mathrsfs}
\usepackage{multirow}

\Title{Surfing the modeling of {\sc pos} taggers in low-resource scenarios}

\TitleCitation{Surfing the modeling of {\sc pos} taggers in low
  resources scenarios}

\Author{Manuel Vilares Ferro $^{1,\dagger,\ddagger}$*\orcidA{},
  V\'{\i}ctor M. Darriba Bilbao $^{2,\ddagger}$, Francisco
  J. Ribadas Pena $^{3,\ddagger}$ and Jorge Gra\~na Gil $^{4,\ddagger}$}

\AuthorNames{Manuel Vilares Ferro, V\'{\i}ctor M. Darriba Bilbao, 
  Francisco J. Ribadas Pena and Jorge Gra\~na Gil}

\AuthorCitation{Vilares, M.; Darriba, V.M.; Ribadas, F.J. and J. Gra\~na.}

\address{%
$^{1}$ \quad Department of Computer Science, University of Vigo; vilares@uvigo.es \\
$^{2}$ \quad Department of Computer Science, University of Vigo; darriba@uvigo.es \\
$^{3}$ \quad Department of Computer Science, University of Vigo; ribadas@uvigo.es \\
$^{4}$ \quad Department of Computer Science, University of A Coru\~na; jorge.grana@udc.es}

\corres{Correspondence: vilares@uvigo.es; (M.V.F.)}

\firstnote{Current address: Edificio Polit\'ecnico, Campus As Lagoas s/n,
  32004 Ourense, Spain} 
\secondnote{These authors contributed equally to this work.}

\abstract{The recent trend towards the application of deep structured
  techniques has revealed the limits of huge models in natural
  language processing. This has reawakened the interest in traditional
  machine learning algorithms, which have proved still to be
  competitive in certain contexts, in particular low-resource
  settings. In parallel, model selection has become an essential task
  to boost performance at reasonable cost, even more so when we talk
  about processes involving domains where the training and/or
  computational resources are scarce. Against this backdrop, we
  evaluate the early estimation of learning curves as a practical
  mechanism for selecting the most appropriate model in scenarios
  characterized by the use of non-deep learners in resource-lean
  settings. On the basis of a formal approximation model
  previously evaluated under conditions of wide availability of
  training and validation resources, we study the reliability of such
  an approach in a different and much more demanding operational
  environment. Using as case study the generation of {\sc
    pos} taggers for Galician, a language belonging to the Western
  Ibero-Romance group, the experimental results are consistent with
  our expectations.}

\keyword{Learning curves; low-resource scenarios; non-deep machine
  learning; model selection; {\sc pos} taggers; stopping criteria}

\newcommand{\abs}[1]{\mid #1 \mid}
\newcommand{\sentences}[1]{\left\llbracket #1 \right\rrbracket}

\newcommand{\sentencesw}[1]{\llceil #1 \rrceil}

\newcommand{\absd}[1]{\left\Vert #1 \right\Vert}

\newcommand{\dinfty}[0]{\pmb{\pmb{\infty}}}

\begin{document}


\section{Introduction}

The application of \textit{machine learning} ({\sc ml}) techniques has
significantly changed the landscapes of \textit{natural language
  processing} ({\sc nlp}) over the last decade, allowing some of the
gaps derived from the use of rule-based approaches to be filled in,
mainly its lack of flexibility and high development cost. Thus, although
the state-of-the-art makes clear that hand-crafted tools are not only
easier to interpret and manipulate but also often provide better
results~\cite{Chiche-Yitagesu-2022,Darwish-etal-2017,Pylypenko-etal-2021,Tayyar-Madabushi-Lee-2016,Zhang-etal-2015},
the high level of dependency on expert knowledge makes their
implementation and subsequent maintenance costly in human terms, in
addition to hindering their applicability to different
languages~\cite{Chiong-Wei-2006,Darwish-etal-2017,Juae-etal-2019,Li-etal-2014,Zhang-etal-2015}. Combined with the surge in computational
power, the possibility of accessing massive amounts of data and the
decline in the cost of disk storage, this has decisively contributed
to the growing popularity of {\sc ml} algorithms as the basis for
classification~\cite{Cramer-2008} and clustering~\cite{Vlachos-2011}
models in a variety of tasks. This includes entity
detection~\cite{Florian2004}, information
retrieval~\cite{Xue:2008:TPC:1390334.1390441}, language
identification~\cite{Chan-etal-2017}, machine
translation~\cite{Libovicky-Helcl-2018}, question
answering~\cite{Cortes-etal-2020}, semantic role
labeling~\cite{Swier-Stevenson-2004}, sentiment
analysis~\cite{Glorot11} and text
classification~\cite{Dai:2007:TNB:1619645.1619732}, among others.

At this juncture, although recent proposals based on \textit{deep
  learning} ({\sc dl}) have outperformed the traditional {\sc ml}
methods on a variety of operating fronts, the approach has also shown
its limitations. Specifically, there are two main reasons for the
popularity and the arguable superiority of the {\sc dl} solutions:
\textit{end-to-end training} and \textit{feature
  learning}\footnote{While end-to-end training allows the model to learn
  all the steps between the initial input phase and the final output
  result, feature learning offers representability to effectively
  encode the information in the data.}. However, the latter translates
into the inability to handle directly
symbols~\cite{Ebrahimi-etal-2021}. This implies that data must first
be converted to vector representations for input into the model and
then do just the reverse with its output, which leads to a complex
interpretability of models. Other well known challenges are the lack
of theoretical foundation~\cite{Tomaso-etal-2020}, the difficulty in
dealing with the long tail~\cite{Hao-etal-2021,Zhang-etal-2021}, the
ineffectiveness at inference and decision
making~\cite{Hoefler-etal-2021}, and the requirement of large amounts
of data and powerful computing resources that may not be
available. This complex picture looks even worse in the {\sc nlp}
domain~\cite{Hang2017}, particularly when it comes to dealing with
\textit{low-resource scenarios}\footnote{The term refers to languages,
  domains or tasks lacking large corpora and/or manually crafted
  linguistic resources sufficient for building {\sc nlp}
  applications.}~\cite{Hedderich-etal-2021-survey}. On the one hand,
feature-based techniques lead to an imperfect use of the linguistic
knowledge. On the other, the scarcity of training data is not only
problematic \textit{per se}, but also because of its impact on the rest
of trials. So, generating high-quality vector representations remains a
challenge~\cite{Chakrabarty-etal-2020} and the imbalance in the training
samples that start the long tail and bias phenomena is more
likely. Together with the prone to overfitting of {\sc dl}
models~\cite{Geman:1992:NNB:148061.148062}, which can result in poor
predictive power, thereby compromising both inference and decision
making.

This has led to a renewal of interest in reviewing the role of {\sc
  dl} techniques \textit{vs.} traditional {\sc ml} ones in the
development of {\sc nlp}
applications~\cite{Hang2017,Magnini-etal-2020,Pylypenko-etal-2021,Saied-etal-2019,Wang-Manning-2013},
particularly in low-resource
scenarios~\cite{Hedderich-etal-2021-survey}. Special attention has
been given here to sequence labeling
tasks~\cite{Wang-Manning-2013}. These encompass a class of {\sc nlp}
problems that involve the assignment of a categorical label to each
member of a sequence of observed values, and whose output facilitates
downstream applications such as parsing or semantic analysis, so
errors at this stage can lower their
performance~\cite{Song:2012:CSP:2390524.2390661}. Among the most
important, we can highlight named entity
recognition~\cite{Juae-etal-2019,Hoesen-Purwarianti-2020}, multi-word
expression identification~\cite{Saied-etal-2019}, and
morphological~\cite{Chakrabarty-etal-2020} and {\sc pos}
tagging~\cite{Darwish-etal-2017,Khan-etal-2019,Ljubesic-2018,Stankovic-etal-2020,Todi-etal-2018}. It
is precisely in this framework, the generation of {\sc pos} taggers
for low-resource scenarios by means of non-deep {\sc ml}, that we
propose the study of model selection based on the early estimation of
learning curves. With that in mind, we first overview the
state-of-the-art and our contribution in
Section~\ref{section-related-work}. Next,
Section~\ref{section-formal-framework} briefly reports on the
theoretical basis supporting our research. In
Section~\ref{section-testing-frame}, we introduce the testing frame
for the experiments described in Section~\ref{section-experiments} and
later discussed in Section~\ref{section-discussion}. Finally,
Section~\ref{section-conclusions} presents our conclusions and
thoughts for future work.


\section{Related Work and Contribution}
\label{section-related-work}

Model selection based on the estimation of learning curves has been
the subject of ongoing research over recent decades, inspired by the
idea that the loss of predictive power and of training are
correlated~\cite{Murata93}. In the scope of {\sc nlp}, these
techniques have been applied to most commonly researched areas such
as, for example, \textit{machine translation}. Specifically, they have
been used here for assessing the quality
systems~\cite{Bertoldi:2012:ELC:2393015.2393076,Turchi:2008:LPM:1626394.1626399},
optimizing parameter setting~\cite{Koehn:2003:SPT:1073445.1073462},
estimating how much training data is required to achieve a certain
degree of accuracy~\cite{Kolachina:2012:PLC:2390524.2390528} or
evaluating the impact of a concrete set of distortion factors on the
performance~\cite{Birch:2008:PSM:1613715.1613809}. Their popularity is
growing especially in the field of {\em active
  learning}\footnote{Those iterative {\sc ml} strategies that interact
  with the environment in each cycle, selecting for annotation the
  instances which are harder to identify.} ({\sc al})
~\cite{Cohn:1994:IGA:189256.189489}, where we can refer to
applications for information
extraction~\cite{Culotta:2005:RLE:1619410.1619452,Thompson:1999:ALN:645528.657614},
parsing~\cite{Becker:2005:TMA:1642293.1642452,Tang:2002:ALS:1073083.1073105}
and text
classification~\cite{Lewis:1994:SAT:188490.188495,Liere97activelearning,McCallum:1998:EEP:645527.757765,Tong:2002:SVM:944790.944793}. The
same is true for {\sc pos}
tagging~\cite{Dagan95committee-basedsampling,Haertel:2008:ACS:1557690.1557708,Ringger:2007:ALP:1642059.1642075}
and closely related tasks, such as named entity
recognition~\cite{Laws:2008:SCA:1599081.1599140,Shen:2004:MAL:1218955.1219030,Tomanek07anapproach}
or word sense
disambiguation~\cite{ChanN07,Chen:2006:ESB:1220835.1220851,Zhu07activelearning},
always with the aim of reducing the annotation effort. Since {\sc al}
prioritizes the data to be labelled in order to maximize the impact
for training a supervised model, it performs better than other {\sc
  ml} strategies with substantially fewer resources. This justifies
the interest in it as an underlying learning guideline to deal with
low-resource
scenarios~\cite{Baldridge08,Ein-Dor-etal-2020,Liu-etal-2018-learning,Lowell-etal-2019}
and specifically in the area of {\sc pos}
tagging~\cite{Anastasopoulos-etal-2018,Chaudhary-etal-2021,Erdmann-etal-2019,Kim-2020,Ringger-etal-2007,Settles-Craven-2008}.

Focusing on the early estimation of learning curves in {\sc al}, we
can distinguish between {\em
  functional}~\cite{Laws:2008:SCA:1599081.1599140,VilaresDarribaRibadas16}
and {\em
  probabilistic}~\cite{Baker-etal-2018,Domhan-etal-2015,Klein-etal-2017}
proposals, depending on the nature of the halting condition used to
determine the end of the training process from the information
generated in each cycle. As a basic difference, functional strategies
not only permit the calculation of relative and absolute error
(resp. convergence) thresholds~\cite{Vilares-etal-2022}\footnote{An
  error (resp. convergence) threshold measures the difference between
  the real and the estimated learning curves at a finite
  (resp. infinite) approximation time. The absolute or relative character is
  applicable to any type of estimation, referring in the first case to
  the strict difference between the values compared, and to that
  existing between values calculated during the prediction process
  in the second one.}, but they are
also simpler and more robust than techniques based on probabilistic
ones. In particular, by replacing single observations with
distributions, we introduce elements of randomness, and thus
uncertainty. That way, to establish how much data is necessary to
reliably build such distributions is no easy matter, and the same
applies to rare event handling. Being related to the definition of a
sampling strategy on a sufficiently wide range of observations, this
question should be preventable or better dealt with in a functional
frame~\cite{VilaresDarribaVilares20}, even more so when the scarcity
of training resources make it difficult to apply probabilistic
criteria.

In such a context, we face the evaluation of {\sc pos} tagging models
by early estimation of learning curves when working in resource-scarce
settings, to the best of our knowledge a yet unexplored
terrain. Leading on from this and looking for an operational solution,
we focus on {\sc al} scenarios, turning our attention to a functional
view of the issue. With a view to exploring the practicality and
potential of the approach, we address it in a setting used previously
to demonstrate its effectiveness when the availability of resources
for learning is not a problem. That way, we take up both the formal
prediction concept and the testing frame introduced
in~\cite{VilaresDarribaRibadas16}, which also allow us to contrast the
level of efficiency to be expected when the conditions for training
and validation of the generated models are much more restrictive.

\section{The Formal Framework}
\label{section-formal-framework}

Below is a brief review of the theoretical basis underlying our work,
taken from~\cite{VilaresDarribaRibadas16}. From now on, we denote the
real numbers by $\mathbb{R}$ and the natural ones by $\mathbb{N}$,
assuming that $0 \not\in \mathbb{N}$. The order in $\mathbb{N}$ is
also extended to $\pmb{\mathbb{N}} := \mathbb{N} \cup \{\infty,
\dinfty{}\}$, in such a way that $\dinfty{} > \infty > i > 0, \;
\forall i \in \mathbb{N}$. Assuming that a learning curve is a plot of
model learning performance over experience, we focus on accuracy as a
measure of that performance.

\subsection{The notational support}

We start with a sequence of observations calculated from cases
incrementally taken from a training data base, and organized around de
concept of \textit{learning scheme}~\cite{VilaresDarribaRibadas16}.

\begin{Definition}
\label{def-learning-scheme} {\em (Learning scheme)}
Let ${\mathcal D}$ be a training data base,
$\mathcal K \subsetneq \mathcal D$ a set of initial items from
$\mathcal D$, and $\sigma:\mathbb{N} \rightarrow \mathbb{N}$ a
function. We define a {\em learning scheme} for $\mathcal D$ with {\em
  kernel} $\mathcal K$ and {\em step} $\sigma$, as a triple
$\mathcal{D}^{\mathcal{K}}_{\sigma}=[\mathcal{K},\sigma,\{\mathcal
D_i\}_{i \in \mathbb{N}}]$, such that
$\{\mathcal D_i\}_{i \in \mathbb{N}}$ is a cover of $\mathcal{D}$
verifying:
\begin{equation}
{\mathcal D}_1 := {\mathcal K} \mbox{ and }
{\mathcal D}_i := {\mathcal
  D}_{i-1} \cup {\mathcal I}_{i}, \; \mathcal I_i
\subset {\mathcal D} \setminus {\mathcal
  D}_{i-1}, \; \absd{{\mathcal I}_{i}}=\sigma(i), \; \forall i \geq 2
\end{equation}
\noindent with $\absd{{\mathcal I}_{i}}$ the cardinality of
${\mathcal I}_{i}$. We refer to $\mathcal{D}_i$ as the {\em individual
  of level} $i$ {\em for} $\mathcal{D}^{\mathcal{K}}_{\sigma}$.
\end{Definition}

That relates a level $i$ with the position
$\absd{\mathcal D_i}$ in the training data base, determining the
sequence of observations
$\{[x_i, {\mathcal A}_{\dinfty{}}[{\mathcal D}](x_i)], \; x_i :=
\absd{\mathcal D_i} \}_{i \in \mathbb{N}}$, where
${\mathcal A}_{\dinfty{}}[{\mathcal D}](x_i)$ is the accuracy achieved
on such an instance by the learner. Thus, a level determines an
iteration in the adaptive sampling whose learning curve is
${\mathcal A}_{\dinfty{}}[{\mathcal D}]$, whilst ${\mathcal K}$
delimits a portion of ${\mathcal D}$ we believe to be enough to
initiate consistent evaluations of the training. For its part,
$\sigma$ identifies the sampling scheduling.

In order to get a reliable assessement, the weak predictor generated
at each learning cycle is extrapolated according to an
\textit{accuracy pattern}~\cite{VilaresDarribaRibadas16}, which allows
to formally compile a set of properties giving stability to the
estimates and widely accepted as \textit{working hypotheses} by the
state-of-the-art in model
evaluation~\cite{Domhan-etal-2015,Domingo:2002:ASM:593433.593526,Meek:2002:LSM:944790.944798,Mohr-Rijn-2021,Schutze:2006:PTP:1183614.1183709,KATRIN08.335}.

\begin{Definition}
\label{def-accuracy-pattern-fitting} {\em (Accuracy pattern)}
Let $C^\infty_{(0,\infty)}$ be the C-infinity functions in
$\mathbb{R}^{+}$, we say that $\pi: \mathbb{R}^{{+}^{n}} \rightarrow
C^\infty_{(0,\infty)}$ is an {\em accuracy pattern} iff $\pi(a_1,
\dots, a_n)$ is positive definite, upper bounded, concave and strictly
increasing.
\end{Definition}

\begin{figure}[H]
\begin{center}
\begin{tabular}{cc}
\hspace*{-.4cm}
\includegraphics[width=0.51\textwidth]{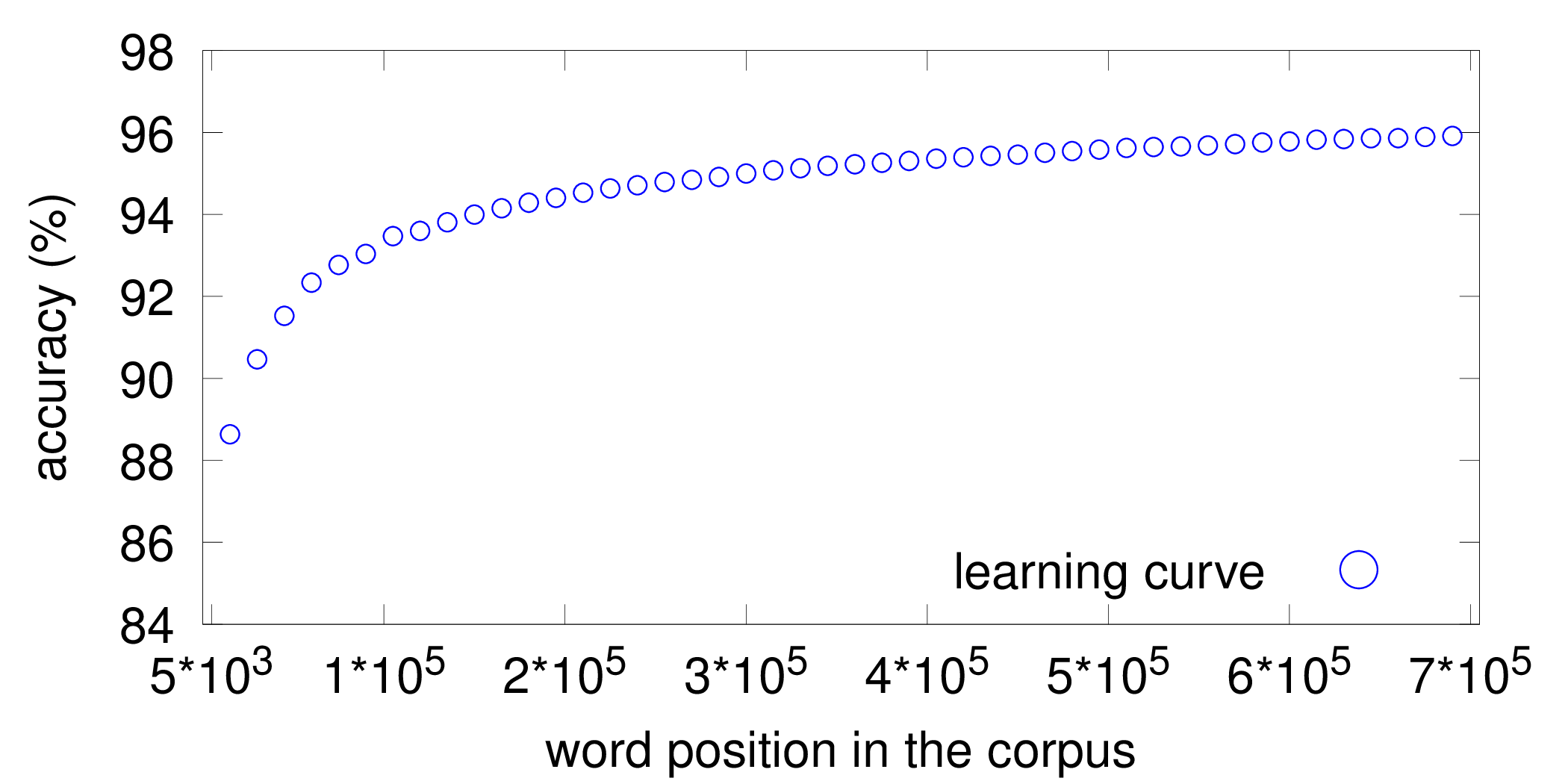} 
&
\hspace*{-.5cm}
\includegraphics[width=0.51\textwidth]{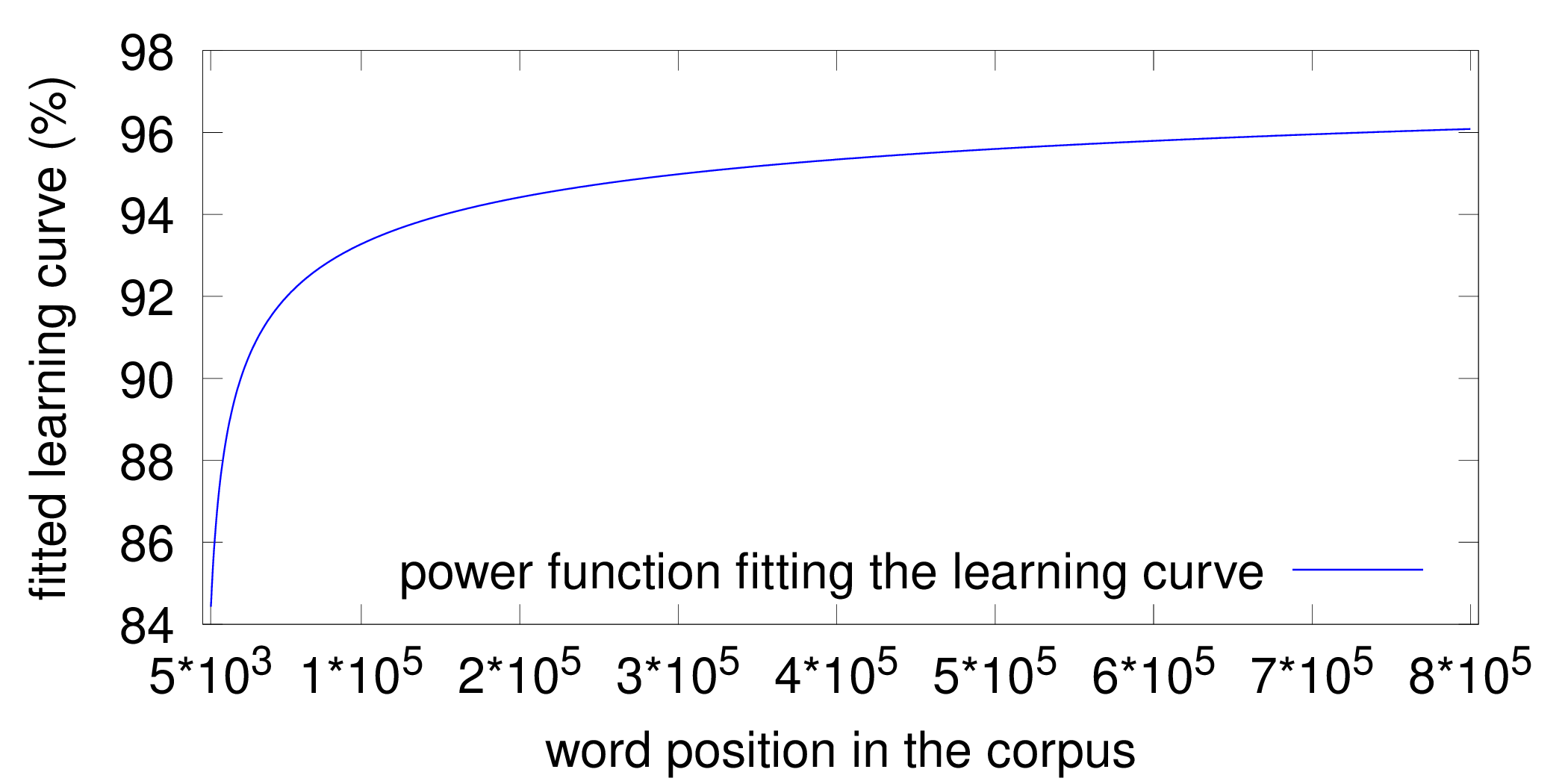}
\end{tabular}
\caption{Learning curve for {\sc svmt}ool on {\sc x}iada, and an
  accuracy pattern fitting it.}
\label{fig-accuracy-SVMTool-XIADA-5000-700000}
\end{center}
\end{figure}

As running accuracy pattern we select the \textit{power law family}
$\pi(a,b,c)(x) :=-a * x^{-b}+c$. Its use is illustrated in the
right-most diagram of
Fig.~\ref{fig-accuracy-SVMTool-XIADA-5000-700000} to fit the learning
curve represented on the left-hand side for the {\sc svmt}ool
tagger~\cite{Gimenez2004} on the {\sc x}iada corpus of
Galician~\cite{XIADAcorpus22}, with the values $a=204.570017$,
$b=0.307277$ and $c=99.226727$ provided by the \textit{trust region
  method}~\cite{Branch1999}. Returning to the review of our notational
support, we now adapt these calculation elements to an iterative
dynamics through the concept of {\em learning
  trend}~\cite{VilaresDarribaRibadas16}.

\begin{Definition}
\label{def-trace} {\em (Learning trend)}
Let $\mathcal{D}^{\mathcal{K}}_{\sigma}$ be a learning scheme, $\pi$
an accuracy pattern and $\ell \in \mathbb{N}, \; \ell \geq 3$ a
position in the training data base $\mathcal{D}$. We define the {\em
learning trend of level} $\ell$ {\em for} ${\mathcal
D}^{\mathcal{K}}_{\sigma}$ {\em using} $\pi$, as a curve ${\mathcal
A}_{\ell}^\pi[{\mathcal D}^{\mathcal{K}}_{\sigma}] \in \pi$, fitting
the observations $\{[x_i, {\mathcal A}_{\dinfty{}}[{\mathcal
D}](x_i)], \; x_i := \absd{\mathcal D_i}
\}_{i=1}^{\ell}$. A sequence of learning trends ${\mathcal
  A}^\pi[{\mathcal D}^{\mathcal {K}}_{\sigma}] :=\{{\mathcal
  A}_{\ell}^\pi[{\mathcal D}^{\mathcal{K}}_{\sigma}]\}_{\ell \in
  \mathbb{N}}$ is called a {\em learning trace}. We refer to
$\{\alpha_\ell\}_{\ell \in \mathbb{N}}$ as the {\em asymptotic
  backbone} of ${\mathcal A}^\pi[{\mathcal D}^{\mathcal
    {K}}_{\sigma}]$, where $y = \alpha_\ell := \lim \limits_{x
  \rightarrow \infty} {\mathcal A}_{\ell}^\pi[{\mathcal
    D}^{\mathcal{K}}_{\sigma}](x)$ is the asymptote of ${\mathcal
  A}_\ell^\pi[{\mathcal D}^{\mathcal {K}}_{\sigma}]$.
\end{Definition}

\begin{figure}[H]
\begin{center}
\begin{tabular}{cc}
\hspace*{-.4cm}
\includegraphics[width=0.51\textwidth]{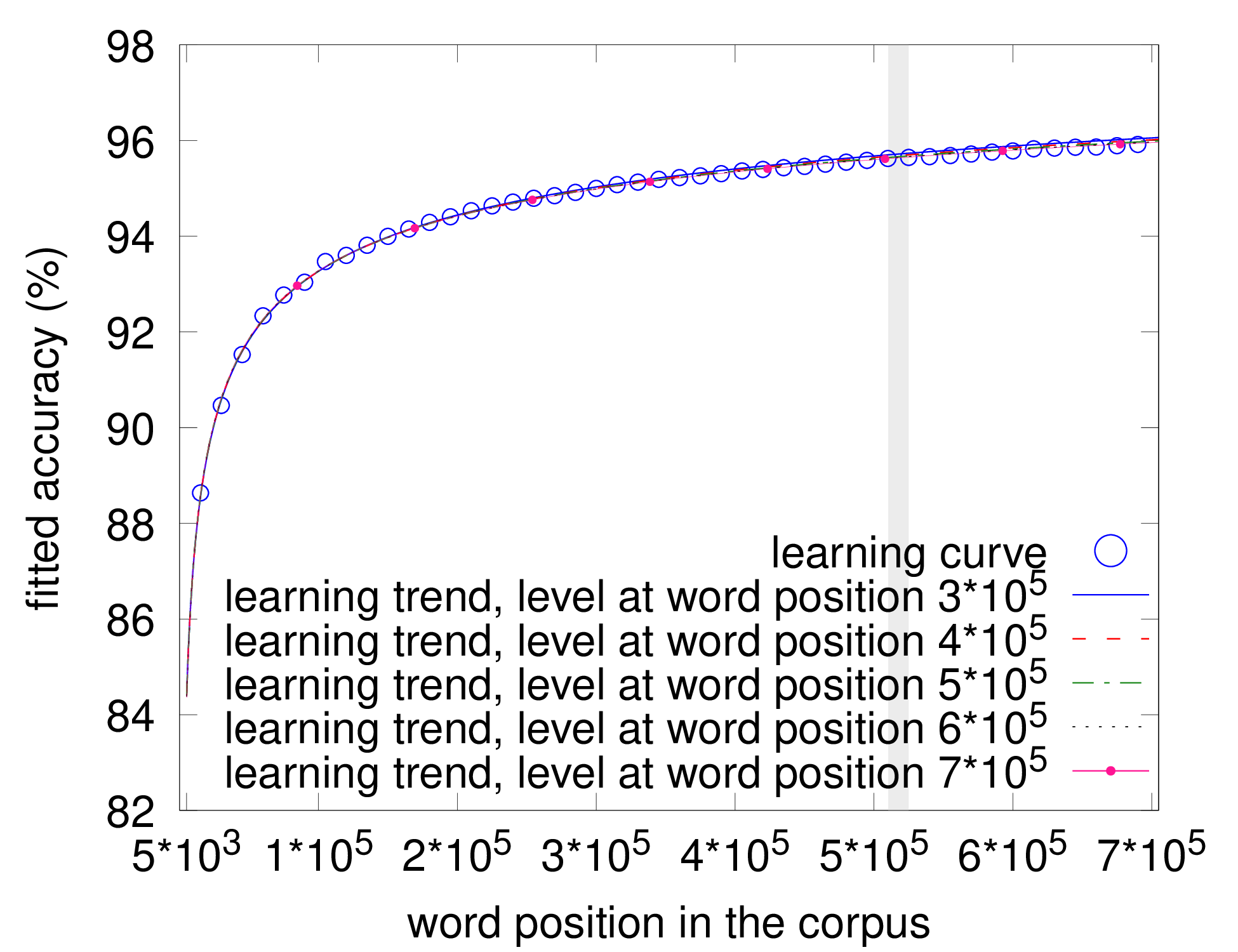}
& 
\hspace*{-.5cm}
\includegraphics[width=0.51\textwidth]{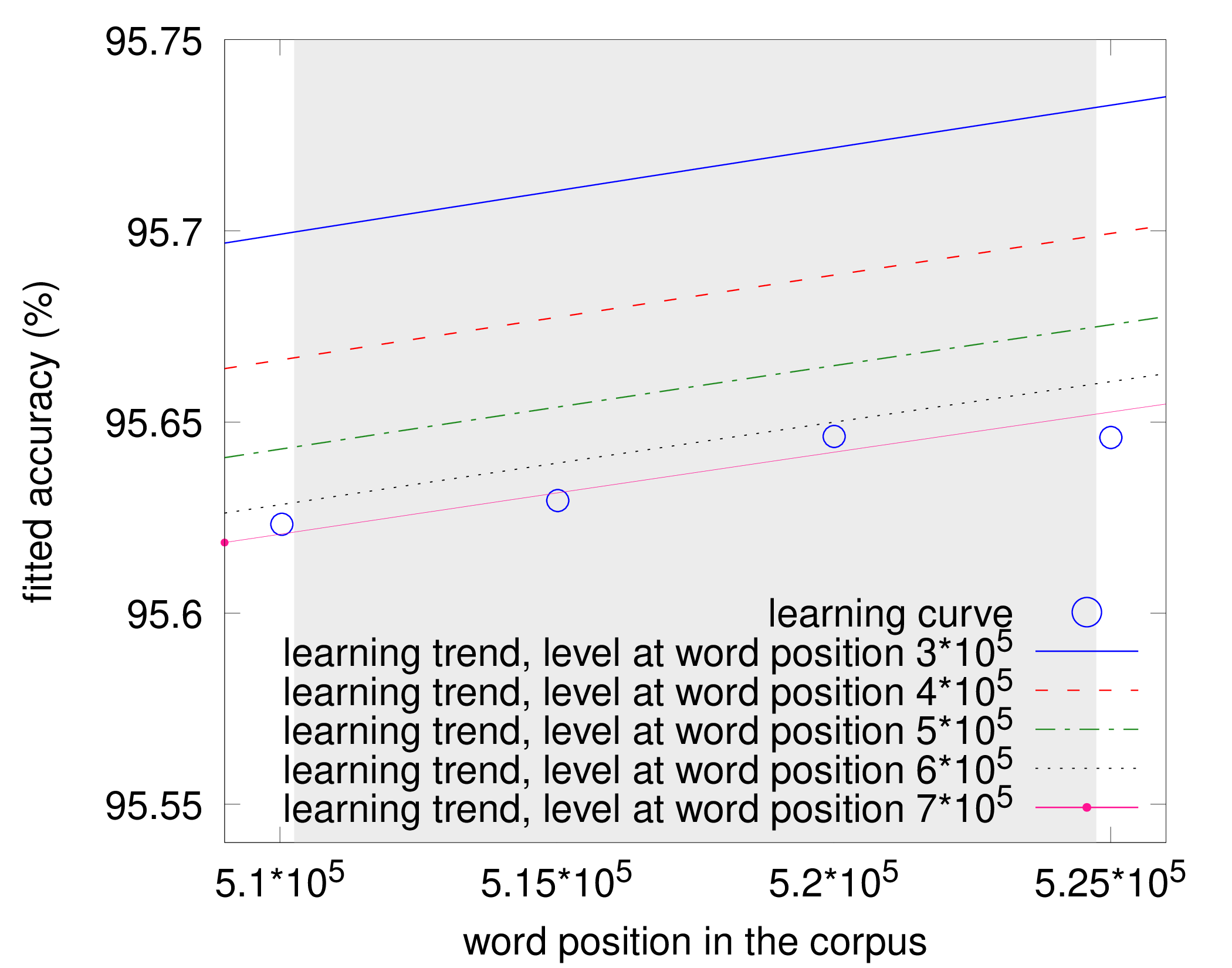}
\end{tabular}
\caption{Learning trace for {\sc svmt}ool on {\sc x}iada, with details
  in zoom.}
\label{fig-trace-and-instantaneous-configuration-nlls-SVMTool-XIADA5000-700000}
\end{center}
\end{figure}

The minimum level $\ell$ for a learning trend ${\mathcal
  A}_{\ell}^\pi[{\mathcal D}^{\mathcal{K}}_{\sigma}]$ is $3$, because
we need at least three points to generate a curve. Its value
${\mathcal A}_{\ell}^\pi[{\mathcal D}^{\mathcal{K}}_{\sigma}](x_i)$ is
our prediction of the accuracy for a case $x_i$, using a model generated
from the first $\ell$ cycles of the learner. Accordingly, the
asymptotic term $\alpha_\ell$ is interpretable as the estimate for the
highest accuracy attainable. Continuing with the tagger {\sc svmt}ool
and the corpus {\sc x}iada,
Fig.~\ref{fig-trace-and-instantaneous-configuration-nlls-SVMTool-XIADA5000-700000}
shows a portion of the learning trace with kernel and uniform
step function $5*10^3$, including a zoom view.

\subsection{Correctness}

Assuming our working hypotheses, the correctness of the proposal --
i.e., the existence and effective approximation of a learning curve
${\mathcal A}_{\dinfty{}}[{\mathcal D}]$ from a subset of its
observations compiled in a learning scheme ${\mathcal D}^{\mathcal
  {K}}_{\sigma}$ -- is demonstrated from the uniform convergence of
the corresponding learning trace ${\mathcal A}^\pi[{\mathcal
    D}^{\mathcal
    {K}}_{\sigma}]$~\cite{VilaresDarribaRibadas16}. Specifically, the
function
\[{\mathcal
  A}_{\infty}^\pi[{\mathcal D}^{\mathcal{K}}_{\sigma}] := {\lim
  \limits_{i \rightarrow \infty}}^u {\mathcal A}_{i}^\pi[{\mathcal
    D}^{\mathcal{K}}_{\sigma}]
\]
\noindent exists and is positive definite, increasing, continuous and
upper bounded by 100 in $(0, \infty)$. In order to estimate the
quality of our approximation, a relative proximity criterion is
introduced. Labelled {\em layered convergence}, it evaluates the
contribution of each learning trend to the convergence process in a
sequence which is proved to be decreasing and convergent to
zero. These layers of convergence can then be interpreted as a reliable
reference to fix error (resp. convergence) thresholds.

\subsection{Robustness}

Robustness is studied from a set of \textit{testing hypotheses}, which
assume that learning curves are positive definite and upper bounded,
albeit only quasi-strictly increasing and concave. An observation
is then categorized according to its position with respect to the {\em
  working level} ({\sc wl}evel), i.e. the cycle after which
irregularities would not impact the correctness. Considering that the
learner stabilizes as the training advances and that the monotonicity of
the asymptotic backbone is at the basis of any halting condition, it
is identified as the level providing the first slope fluctuation below
a given ceiling in such a backbone and, once passed, the
\textit{prediction level} ({\sc pl}evel) marking the beginning of
learning trends which could feasibly predict the learning curve,
namely not exceeding its maximum
(100)~\cite{VilaresDarribaRibadas16}. Based on this, {\sc wl}evel is
calculated as the lowest level for which the normalized absolute value
of the slope of the line joining two consecutive points on the
asymptotic backbone is less than a verticality threshold $\nu$, which
is corrected by applying a coefficient $1/\varsigma$ for avoiding
having to deal with to both infinitely large slopes and extremely small
decimal fractions.

\begin{Definition}
\label{def-level-of-work-trace}{\em (Working and prediction levels)}
Let ${\mathcal A}^\pi[{\mathcal D}^{\mathcal {K}}_{\sigma}]$ be a
learning trace with asymptotic backbone
$\{\alpha_i\}_{i \in \mathbb{N}}$, $\nu \in (0, 1)$,
$\varsigma \in \mathbb{N}$ and $\lambda \in \mathbb{N} \cup
\{0\}$. The {\em working level} {\em (}{\sc wl}{\em evel}{\em)} {\em for}
${\mathcal A}^\pi[{\mathcal D}^{\mathcal {K}}_{\sigma}]$ {\em with
  verticality threshold} $\nu$, {\em slowdown} $\varsigma$ {\em and
  look-ahead} $\lambda$, is the smallest
$\omega(\nu,\varsigma,\lambda) \in \mathbb{N}$ verifying
\begin{equation}
\label{equation-permissible-verticality-trace}
\frac{\sqrt[\varsigma]{\nu}}{1 - \nu} \geq \frac{\abs{\alpha_{i+1} -
    \alpha_{i}}}{x_{i+1} - x_i}, \; x_i :=
\absd{{\mathcal D}_{i}}, \; \forall i \mbox{ such that } 
\omega(\nu,\varsigma,\lambda) \leq i \leq
\omega(\nu,\varsigma,\lambda) + \lambda
\end{equation}
\noindent while the smallest $\wp(\nu,\varsigma,\lambda) \geq
\omega(\nu,\varsigma,\lambda)$ with
$\alpha_{\wp(\nu,\varsigma,\lambda)} \leq 100$ is the {\em prediction
  level} {\em (}{\sc pl}{\em evel}{\em )}. 
\end{Definition}

\begin{figure}[H]
\begin{center}
\includegraphics[width=0.61\textwidth]{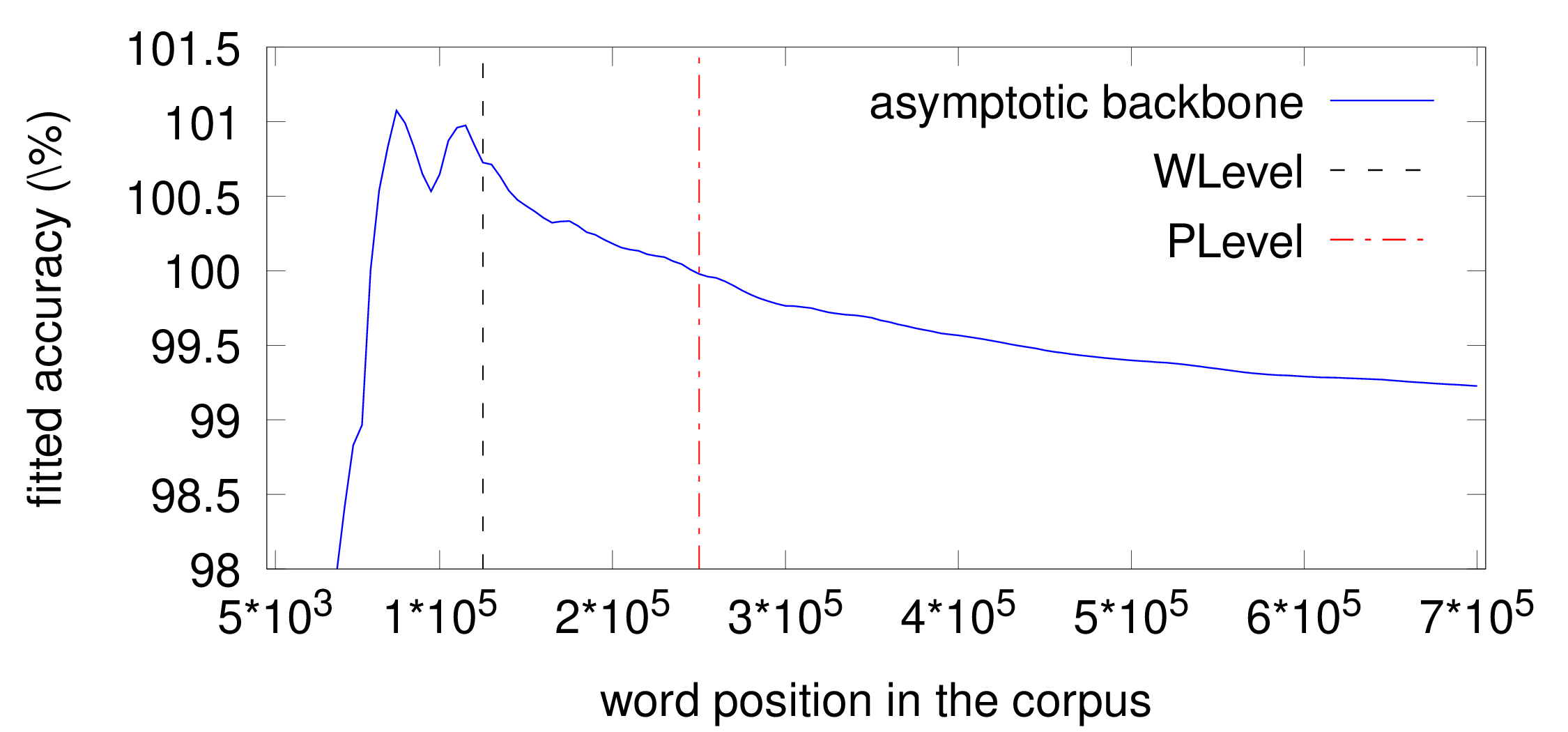} 
\end{center}
\caption{Working and prediction levels for {\sc svmt}ool on {\sc x}iada,
  with details in zoom.}
\label{fig-lashes-trace-and-level-sequence-nlls-SVMTool-XIADA-5000-700000}
\end{figure}

Following our example,
Fig.~\ref{fig-lashes-trace-and-level-sequence-nlls-SVMTool-XIADA-5000-700000}
shows the scale of the deviations in the asymptotic backbone before
and after {\sc wl}evel, for $\nu=2*10^{-5}$, $\varsigma=1$ and
$\lambda=5$. The {\sc pl}evel is also displayed, proving that these two
levels might not be the same.

\section{The Testing Frame}
\label{section-testing-frame}

Given a training corpus $\mathcal{D}$, we want to study how far in
advance and how well a learning curve
$\mathcal{A}_{\dinfty{}}[\mathcal{D}^{\mathcal{K}}_{\sigma}]$, built
from a kernel $\mathcal{K}$ and using a step function $\sigma$, can be
approximated in a low-resource scenario. To ensure the relevance of
the results obtained, we will standardize the conditions under which
the experiments take place, following the same criteria previously
considered in the study of resource-rich
languages~\cite{VilaresDarribaRibadas16}.

\subsection{The monitoring structure}

As evaluation basis we consider the {\em
  run}~\cite{VilaresDarribaRibadas16}, a tuple
$\mathcal{E}=[\mathcal{A}^\pi[\mathcal{D}^{\mathcal{K}}_{\sigma}],
  \wp(\nu,\varsigma,\lambda),\tau]$ characterized by a learning trace
$\mathcal{A}^\pi[\mathcal{D}^{\mathcal{K}}_{\sigma}]$, a prediction
level $\wp(\nu,\varsigma,\lambda)$ and a convergence threshold
$\tau$. We then apply our study on a collection of runs ${\mathcal C}
=\{{\mathcal E}_i\}_{i \in I}$, defined for a set of different
learners. In order to avoid misconceptions due to the lack of
uniformity in the testing frame, a common corpus $\mathcal{D}$, kernel
size, accuracy pattern $\pi$, step function $\sigma$, verticality
threshold $\nu$, slowdown $\varsigma$, look-ahead $\lambda$ and
convergence threshold $\tau$ are used.

In practice, we are interested in studying each run
$\mathcal{E}=[\mathcal{A}^\pi[\mathcal{D}^{\mathcal{K}}_{\sigma}],
  \wp(\nu,\varsigma,\lambda),\tau]$ from the level in which
predictions are below $\tau$, and which we baptize \textit{convergence
  level} ({\sc cl}evel). So, once the {\sc pl}evel is found during the
computation of the trace
$\mathcal{A}^\pi[\mathcal{D}^{\mathcal{K}}_{\sigma}]$, we begin to
check the layer of convergence. When it reaches the threshold $\tau$,
the trend $\mathcal{A}_{\mbox{\footnotesize {\sc
      cl}evel}}^\pi[\mathcal{D}^{\mathcal{K}}_{\sigma}]$ becomes the
model for the learning curve
$\mathcal{A}_{\dinfty{}}[\mathcal{D}^{\mathcal{K}}_{\sigma}]$, and the
process of approximation is stopped.

For the runs ${\mathcal C} =\{{\mathcal E}_i\}_{i \in I}$,
monitoring is applied to the learning trends
$\{\mathcal{A}_{\mbox{\footnotesize {\sc
      cl}evel}_i}^\pi[\mathcal{D}^{\mathcal{K}}_{\sigma}]\}_{i \in I}$
on a finite common {\em control sequence} of levels for the training
data base, which are extracted from an interval of the
\textit{prediction windows} $\{[\mbox{{\sc cl}evel}_i, \infty)\}_{i
    \in I}$~\cite{VilaresDarribaRibadas16}. In each {\em control
    level}, the {\em accuracy} (Ac) and the corresponding {\em
    estimated accuracy} (EAc) are computed for each run using six
  decimal digits, though only two are represented for reasons of space
  and visibility.

\subsection{The performance metrics}

Our aim is both to assess the reliability of our estimates and their
robustness against variations in the working hypotheses. To do so, we
employ two specific kind of metrics~\cite{VilaresDarribaRibadas16}.

\subsubsection{Measuring the reliability}

We here differentiate two complementary viewpoints: quantitative
and qualitative. In the first case, it is simply a matter of studying
the closeness of the estimates and the actual learning curves, while in
the second the objective is to determine the impact of those
estimates on the decision making about the performance of some models
relative to others.

\paragraph{The quantitative perspective}

A simple way of measuring the reliability from this viewpoint is
through the {\em mean absolute percent error} ({\sc
  mape})~\cite{Vandome63}. For every run $\mathcal{E}$ and level $i$
of a control sequence $\mathcal{S}$, we first compute the
\textit{percentage error} ({\sc pe}) as the difference between the EAc
calculated from $\mathcal{A}_{\mbox{\footnotesize {\sc
      cl}evel}_{\mathcal
    E}}^\pi[\mathcal{D}^{\mathcal{K}}_{\sigma}](i)$ and the Ac from
$\mathcal{A}_{\dinfty{}}[\mathcal{D}^{\mathcal{K}}_{\sigma}](i)$. We
can then express the {\sc mape} as the arithmetic mean of the unsigned
{\sc pe}~\cite{VilaresDarribaRibadas16}, as
\begin{equation}
\mbox{\sc pe}(\mathcal{E})(i) := 100 *
\frac{[\mathcal{A}_{\mbox{\footnotesize
    {\sc cl}evel}_{\mathcal E}}^\pi -
      \mathcal{A}_{\dinfty{}}][\mathcal{D}^{\mathcal{K}}_{\sigma}](i)}
     {\mathcal{A}_{\dinfty{}}[\mathcal{D}^{\mathcal{K}}_{\sigma}](i)}, \;
     \mathcal{E} = [{\mathcal A}^\pi[{\mathcal D}^{\mathcal{K}}_{\sigma}],
                    \wp(\nu,\varsigma,\lambda), \tau],
     \; i \in {\mathcal S}
\end{equation}
\begin{equation}
\mbox{\sc mape}(\mathcal{E})({\mathcal S}) := 
\frac{100}{\absd{\mathcal S}} * 
\sum_{i \in {\mathcal S}} \abs{\mbox{\sc pe}(\mathcal{E})(i)}
\end{equation}

\noindent Intuitively, the error in the estimates done over a control
sequence is, on average, proportional to the {\sc mape}, who fulfil
our requirements at this point.

\paragraph{The qualitative perspective}

To that end, having fixed a collection of runs ${\mathcal H}$ working
on a common corpus and a control sequence ${\mathcal S}$, the
reliability of one of such runs depends on the percentage of cases on
which its estimates not altering the relative position of its learning
curve with respect to the rest throughout ${\mathcal S}$. In this
sense, our primary reference is the \textit{reliability estimation}
({\sc re}) \textit{of two runs} ${\mathcal E},\tilde{\mathcal E} \in
{\mathcal H}$ \textit{on} $i \in {\mathcal S}$,
defined~\cite{VilaresDarribaRibadas16} as
\begin{equation}
\label{equation-re-intersection}
\begin{array}{c}
\mbox{{\sc re}}(\mathcal{E},\tilde{\mathcal{E}})(i) :=
\left\{ \begin{array}{l}
          1 \mbox{ if } \; [[\mathcal{A}_{\dinfty{}}-
                            \tilde{\mathcal{A}}_{\dinfty{}}]
              * [\mathcal{A}_{\mbox{\footnotesize {\sc cl}evel}_{\mathcal E}}^\pi -
                    \tilde{\mathcal{A}}_{\mbox{\footnotesize
    {\sc cl}evel}_{\tilde{\mathcal E}}}^\pi]][\mathcal{D}^{\mathcal{K}}_{\sigma}](i)
              \geq 0 \\
          0 \mbox{ otherwise}
        \end{array}
\right.
\end{array}
\end{equation}
\noindent with $\mathcal{E} = [{\mathcal A}^\pi[{\mathcal
      D}^{\mathcal{K}}_{\sigma}], \wp(\nu,\varsigma,\lambda), \tau]$,
$\tilde{\mathcal E} = [\tilde{\mathcal A}^\pi[{\mathcal
      D}^{\tilde{\mathcal K}}_{\tilde{\sigma}}],
  \wp(\nu,\varsigma,\lambda), \tilde\tau]$ and $\mathcal{E} \neq
\tilde{\mathcal E}$. Having fixed a control level, this Boolean
function verifies if the estimates for ${\mathcal E}$ and
$\tilde{\mathcal E}$ preserve the relative positions of the
corresponding observations, and it can be extended to the control
sequence through the concept of {\em reliability estimation
  ratio}~\cite{VilaresDarribaRibadas16}.

\begin{Definition}
\label{def-reliability-estimation-ratio} {\em (Reliability estimation
  ratio)} Let ${\mathcal E}$ and $\tilde{\mathcal {E}}$ runs on a
control sequence ${\mathcal S}$. We define the {\em reliability
  estimation ratio} {\sc (rer)} {\em of} ${\mathcal E}$ {\em and}
$\tilde{\mathcal E}$ {\em for} ${\mathcal S}$ as
\begin{equation}
\label{equation-rer}
\mbox{\sc rer}(\mathcal{E},\tilde{\mathcal{E}})({\mathcal S}) := 100 *
\frac{\sum_{i \in {\mathcal S}} \mbox{\sc re}(\mathcal{E},\tilde{\mathcal{E}})(i)}{\absd{{\mathcal S}}}
\end{equation}
\end{Definition}

From this, we can calculate the percentage of runs in a set ${\mathcal
  H}$ with regard to which the estimates for a given one $\mathcal{E}$
are reliable on the whole of the control sequence $\mathcal{S}$
considered. We denote the resulting metric as {\em decision-making
  reliability}~\cite{VilaresDarribaRibadas16}.

\begin{Definition}
\label{def-decision-making-reliability} {\em (Decision-making
  reliability)} Let ${\mathcal H} =\{{\mathcal E}_k\}_{k \in K}$ and
${\mathcal E} \not\in {\mathcal H}$ a set of runs and a run,
respectively, on a control sequence ${\mathcal S}$. We define the {\em
  decision-making reliability ({\sc dmr}) \em of} ${\mathcal E}$ {\em
  on} ${\mathcal H}$ {\em for} ${\mathcal S}$ as
\begin{equation}
\label{equation-dmr}
\mbox{\sc dmr}(\mathcal{E},\mathcal{H})({\mathcal S}) := 100 *
\frac{\absd{\mathcal{E}_k \in \mathcal{H}, \; \mbox{{\sc
        rer}}(\mathcal{E},\mathcal{E}_k)({\mathcal S}) = 100}} 
     {\absd{\mathcal{S}}}
\end{equation}
\end{Definition}

\subsubsection{Measuring the robustness}

Since the stability of a run ${\mathcal E}$ correlates to the degree
of monotony in its asymptotic backbone, we measure it as the
percentage of monotonic elements in the latter through the interval
$[\mbox{{\sc wl}evel}_{\mathcal E}, \mbox{{\sc cl}evel}_{\mathcal E}]$
where the approximation performs effectively. We baptize it as {\em
  robustness rate}~\cite{VilaresDarribaRibadas16}.

\begin{Definition}
\label{def-robustness-rate} {\em (Robustness rate)}
Let ${\mathcal E}$ be a run with asymptotic backbone
$\{\alpha_\ell\}_{\ell \in \mathbb{N}}$, and $\mbox{\em {\sc
    cl}evel}_{\mathcal E}$ and $\mbox{\em {\sc wl}evel}_{\mathcal E}$
its convergence and working levels, respectively. We define the {\em
  robustness rate ({\sc rr}) of} ${\mathcal E}$ as 
\begin{equation}
\label{equation-rr}
\mbox{\sc rr}(\mathcal{E}) := 100 *
\frac{\absd{\mu}}
     {\absd{\{\alpha_i, \; \mbox{\em {\sc wl}evel}_{\mathcal E} \leq i \leq
\mbox{\em {\sc cl}evel}_{\mathcal E}\}}}
\end{equation}

\noindent with $\mu$ the longest maximum monotonic subsequence of
$\{\alpha_i, \; \mbox{\em {\sc wl}evel}_{\mathcal E} \leq i \leq
\mbox{\em {\sc cl}evel}_{\mathcal E}\}$.
\end{Definition}

The tolerance of a run to variations in the working hypotheses is
therefore greater the higher its {\sc rr}, thus providing a simple
criterion for checking the degree of robustness on which we can count.


\section{The Experiments}
\label{section-experiments}

Within a model selection context and focused on the generation of {\sc
  pos} taggers in low-resource scenarios, our goal is to provide
evidence of the suitability of using evaluation mechanisms based on the
early estimation of learning curves. It is a non trivial challenge
because, to provide that evidence, we need to study our estimates over
a significant range of observations, which is in clear
contradiction to the scarcity of training data.

\subsection{The linguistics resources}

In order to address the issue posed, we chose to work with a case
study that meets four conditions. The first is that the language
considered is really a resource-poor one, which should guarantee that
it has been outside the tuning phase of the process of developing
nthe taggers later used in the experiments, thereby precluding any
potential biases associated with the learning architecture. Second, it
should have a rich morphology, thus making the training process
non-trivial and therefore relevant to the test performed. Thirdly, we
should have at least a training corpus of sufficient size to study the
reliability of the results obtained. Finally, that corpus should
provide sufficiently low levels of convergence to allow the
identification of the learning processes with the generation of viable
models from a small set of training data.

We then take as a case study Galician, a member of the West Iberian
group of Romance languages that also includes the better-known
Portuguese. It is an inflectional language with a great variety of
morphological processes, particularly non-concatenative ones, derived
from its Latin origin. Some of its most distinctive characteristics
are~\cite{VilGraAraCabDiz98a}:

\begin{itemize}

\item A highly complex conjugation paradigm, with 10 simple tenses
      including the Infinitive conjugate, all of which have 6
      different persons. If we add the Present Imperative with 2
      forms, non-conjugated Infinitive, Gerund and Participle, then 65
      inflected forms are associated with each verb.

\item Irregularities in both verb stems and endings. Common verbs,
  such as \verb$facer$ ({\em to do}), have up to 5 stems:
  \verb$fac-er$, \verb$fag-o$, \verb$fa-s$, \verb$fac-emos$,
  \verb$fix-en$. Approximately 30\% of verbs are irregular.
      
\item Verbal forms with enclitic pronouns at the end, which can produce
  changes in the stem due to the presence of accents: \verb$deu$ ({\em
    gave}), {\tt d\'{e}ullelo} ({\em he/she gave it to them}). The
  unstressed pronouns are usually suffixed and, moreover, they can be
  easily drawn together and often are contracted ({\tt lle + o} = {\tt
    llo}), as in the case of {\tt v\'{a}itemello buscar} ({\em go and
    fetch it for him (do it for me)}). It is also frequent to use
  what we call a {\em solidarity pronoun}, as {\tt che} and {\tt vos},
  in order to let the listeners be participant in the action. That
  way, forms with up to four enclitic pronouns, like {\tt
    perd\'{e}uchellevolo} ({\em he had lost it to him}), are rather
  common.

\item A highly complex gender inflection, including words with only one
  gender as \verb$home$ ({\em man}) and \verb$muller$ ({\em woman}),
  and words with the same form for both genders as \verb$azul$ ({\em
    blue}). Regarding words with separate forms for masculine and
  feminine, more than 30 variation groups are identified.
      
\item A highly complex number inflexion, with words only being
  presented in singular form, such as {\tt luns} ({\em monday}), and
  others where only the plural form is correct, as {\tt matem\'aticas}
  ({\em mathematics}). More than a dozen variation groups are 
  identified.

\end{itemize}

\noindent This choice limits the availability of curated corpora of
sufficient size to a single candidate, {\sc
  x}iada~\cite{XIADAcorpus22}, whose latest version (2.8) includes
over 747,000 entries gathered from three different sources: general
and economic news articles, and short stories. With the aim of
accommodating the elaborate linguistic structure previously described,
the tag-set includes 460 tags, a short description of which can be
found at {\em http://corpus.cirp.gal/xiada/etiquetario/taboa}.

\subsection{The {\sc pos} tagging systems}

As already argued, we focus on models built from {\sc al}, selecting a
broad range of proposals covering the most representative non-deep
learning architectures\footnote{Our reference here is the
  state-of-the-art in {\sc pos} tagging by the {\em Association for
  Computational Linguistics} ({\sc acl}), available at the link {\tt
    https://aclweb.org/aclwiki/index.php?title=POS\_Tagging\_(State\_of\_the\_art)}.},
the same tested in~\cite{VilaresDarribaRibadas16} on resource-rich
languages. This matching will allow us to establish, together with the
subsequent identification of the parameters in the testing space, a
valid reference based on the results obtained in that work:

\begin{itemize}

\item In the category of stochastic methods and representing the
  \textit{hidden M\'arkov models} ({\sc hmm}s), we chose {\sc t}n{\sc
    t}~\cite{Brants2000}. We also include the {\sc t}ree{\sc
    t}agger~\cite{Schmid1994}, a proposal that uses decision trees to
  generate the {\sc hmm}, and {\sc m}orfette~\cite{Chrupala2008}, an
  averaged perceptron approach~\cite{Collins2002}. To illustrate the
  \textit{maximum entropy models} ({\sc mem}s), we select {\sc
    mxpost}~\cite{Ratnaparki1996} and {\sc o}pen{\sc nlp} {\sc
    m}ax{\sc e}nt~\cite{Toutanova2003}. Finally, the
  \textsc{s}tanford {\sc pos} tagger~\cite{Toutanova2003} combines
  features of {\sc hmm}s and {\sc mem}s using a \textit{conditional
    M\'arkov model}.

\item Under the heading of other approaches we consider fn{\sc
    tbl}~\cite{Ngai2001}, an update of the classic {\sc
    b}rill tagger~\cite{Brill1995a}, as an example of
  transformation-based learning. As memory-based method we take the
  \textit{memory-based tagger} ({\sc mbt})~\cite{Daelemans1996a},
  while {\sc svmt}ool~\cite{Gimenez2004} illustrates the behaviour
  of \textit{support vector machines} ({\sc svm}s).
\end{itemize}

\noindent In addition, this ensures an adequate coverage of the range
of learners available in the computational domain under consideration.

\subsection{The testing space}

Following the way drawn by the choice of {\sc ml} architectures
discussed above, the design of the testing space will be the same as
the one considered in the
state-of-the-art~\cite{VilaresDarribaRibadas16} for the study of
resource-rich languages, thus ensuring the reference value of the
latter. Thus, in order to avoid dysfunctions resulting from
sentence truncation during training, we retake the class of learning
scheme then proposed, which permits us to reap the maximum from the
training process. Given a corpus $\mathcal D$, a kernel $\mathcal K
\subsetneq \mathcal D$ and a step function $\sigma$, we build the set
of individuals $\{\mathcal D_i\}_{i \in \mathbb{N}}$ as follows:
\begin{equation}
\begin{array}{l}
\hspace*{-.15cm} {\mathcal D}_i := \sentences{{\mathcal C}_i}, \mbox{ with } {\mathcal C}_1 := {\mathcal K} \mbox{
  and } {\mathcal C}_i := {\mathcal C}_{i-1} \cup {\mathcal I}_{i}, \;
\mathcal I_i \subset {\mathcal C} \setminus {\mathcal C}_{i-1}, \;
\absd{{\mathcal I}_{i}} := \sigma(i), \; \forall i \geq 2
\end{array}
\end{equation}
\noindent where $\sentences{{\mathcal C}_{i}}$ denotes the minimal set of
sentences including ${\mathcal C}_{i}$.

Along the same lines and with respect to the setting of runs, the size
of the kernels is $5*10^3$ words and the constant step function
$5*10^3$ locates the instances, which can be considered conservative
values since smaller and larger ones are possible. Regarding the
parameters used for estimating the prediction levels, the choice again
goes to $\nu=4*10^{-5}$, $\varsigma=1$ and $\lambda=5$. This also
holds true for the selection of the regresion technique used for
approximating the partial learning curves and for $\pi$, that falls on
the \textit{trust region method}~\cite{Branch1999} and the power law
family~\cite{Gu:2001:MCP:645940.671380}, respectively.

Taking into account that real corpora are finite, we study the
prediction model within their boundaries, which implies limiting the
scope in measuring the layers of convergence. We then adapt 
the sampling window and the control levels to the size of the corpus
now considered. So, if $\sentencesw{\ell}$ denotes the position of the
first sentence-ending beyond the $\ell$-th word, the former comprises
the interval $[\sentencesw{5*10^3}, \sentencesw{7*10^5}]$, whilst the
latter are taken from control sequences in $[\sentencesw{3*10^5},
  \sentencesw{7*10^5}]$. In order to confer additional
stability on our measures, we apply a $k$-fold cross
validation~\cite{Clark2010} to compute the samples, with $k=10$.


\begin{table}[H]
\caption{Monitoring of runs along the control sequences.}
\label{table-statistics-on-accuracy-no-anchor-5000-700000}
\begin{center}
\begin{scriptsize}
\begin{tabular}{@{\hspace{0pt}}l@{\hspace{3pt}}l@{\hspace{2pt}}
                 r@{\hspace{0pt}}c@{\hspace{2pt}}  
                 c@{\hspace{0pt}}c@{\hspace{2pt}}  
                 r@{\hspace{0pt}}c@{\hspace{2pt}}  
                 c@{\hspace{4pt}}c@{\hspace{3pt}}c@{\hspace{4pt}}|@{\hspace{4pt}}                  
                                 c@{\hspace{3pt}}c@{\hspace{3pt}}|@{\hspace{2pt}}                  
                                 c@{\hspace{3pt}}c@{\hspace{3pt}}|@{\hspace{2pt}}                  
                                 c@{\hspace{3pt}}c@{\hspace{3pt}}|@{\hspace{4pt}}                  
                                 c@{\hspace{3pt}}c@{\hspace{3pt}}c@{\hspace{2pt}}                  
                 c@{\hspace{0pt}}c@{\hspace{2pt}}r@{\hspace{0pt}}c@{\hspace{2pt}}r@{\hspace{0pt}}} 

 & & $\mbox{\bf{\scshape pl}evel}^{\tiny \mbox{\bf wp}}$ & & \boldmath$\tau$ & & $\mbox{\bf{\scshape cl}evel}^{\tiny \mbox{\bf wp}}$ & & \multicolumn{11}{c}{$\mbox{\bf Control-Level}^{\tiny \mbox{\bf wp}}$} & & \bf\scshape mape & & \bf\scshape dmr~ & & \multicolumn{1}{c}{\bf\scshape rr} \\
\cline{1-1} \cline{3-3} \cline{5-5} \cline{7-7} \cline{9-19} \cline{21-21} \cline{23-23} \cline{25-25}

& & & & & & & & \multicolumn{3}{c}{$\sentencesw{3*10^5}$} & \multicolumn{2}{c}{$\sentencesw{4*10^5}$} & \multicolumn{2}{c}{$\sentencesw{5*10^5}$} & \multicolumn{2}{c}{$\sentencesw{6*10^5}$} & \multicolumn{2}{c}{$\sentencesw{7*10^5}$} & & & & & & \rule{0pt}{2.5ex} \\
\cline{1-1} \cline{3-3} \cline{5-5} \cline{7-7} \cline{9-19} \cline{21-21} \cline{23-23} \cline{25-25}

& & & & & & & & \multicolumn{2}{c}{\bf Ac} & \bf EAc & \bf Ac & \bf EAc & \bf Ac & \bf EAc & \bf Ac & \bf EAc & \bf Ac & \bf EAc & 
 & & & & & \rule{0pt}{3ex} \\
\cline{1-1} \cline{3-3} \cline{5-5} \cline{7-7} \cline{9-19} \cline{21-21} \cline{23-23} \cline{25-25}

fn\textsc{tbl} & & 105.003~~ & & 2.40 & & 150.017~~ & & & 94.16 & 93.87 & 94.57 & 94.30 & 94.96 & 94.61 & 95.16 & 94.84 & 95.34 & 95.03 & & 0.32 & & 85.71 & \rule{0pt}{4ex} & 90.00 \\
\textsc{m}ax\textsc{e}nt & & 110.047~~ & & 2.50 & & 135.019~~ & & & 92.90 & 92.78 & 93.30 & 93.19 & 93.58 & 93.48 & 93.85 & 93.70 & 94.08 & 93.88 & & 0.15 & & 100.00 & & 100.00 \\
\textsc{mbt} & & 85.012~~ & & 2.20 & & 145.016~~ & & & 92.97 & 92.84 & 93.42 & 93.22 & 93.76 & 93.50 & 94.01 & 93.72 & 94.30 & 93.89 & & 0.28 & & 100.00 & & 92.31 \\
\textsc{m}orfette & & 75.011~~ & & 2.60 & & 105.003~~ & & & 94.61 & 94.54 & 94.98 & 94.89 & 95.21 & 95.14 & 95.41 & 95.33 & 95.55 & 95.49 & & 0.09 & & 100.00 & & 85.71 \\
\textsc{mxpost} & & 110.047~~ & & 2.30 & & 145.016~~ & & & 93.44 & 93.17 & 93.88 & 93.57 & 94.20 & 93.85 & 94.44 & 94.06 & 94.63 & 94.23 & & 0.35 & & 100.00 & & 100.00 \\
\textsc{s}tanford & & 95.015~~ & & 2.40 & & 125.001~~ & & & 94.41 & 94.43 & 94.78 & 94.80 & 95.07 & 95.07 & 95.26 & 95.27 & 95.41 & 95.43 & & 0.02 & & 85.71 & & 85.71 \\
\textsc{svmt}ool & & 250.012~~ & & 2.20 & & 250.012~~ & & & 95.00 & 95.05 & 95.36 & 95.44 & 95.60 & 95.71 & 95.78 & 95.93 & 95.92 & 96.10 & & 0.12 & & 100.00 & & 86.67 \\
\textsc{t}n\textsc{t} & & 85.012~~ & & 2.00 & & 130.003~~ & & & 94.47 & 94.38 & 94.79 & 94.70 & 95.05 & 94.93 & 95.23 & 95.10 & 95.35 & 95.23 & & 0.12 & & 71.43 & & 100.00 \\
\textsc{t}ree\textsc{t}agger & & --\hspace{12pt} & & 2.10 & & --\hspace{12pt} & & & 93.36 & -- & 93.77 & -- & 94.02 & -- & 94.28 & -- & 94.42 & -- & & -- & & --\hspace{8pt} & & --\hspace{8pt} \\

\cline{1-1} \cline{3-3} \cline{5-5} \cline{7-7} \cline{9-19} \cline{21-21} \cline{23-23} \cline{25-25}
\end{tabular}
\end{scriptsize}
\end{center}
\end{table}

\section{Discussion}
\label{section-discussion}

As mentioned, the experiments are studied from two complementary
points of view, quantitative and qualitative, according to the 
performance metrics previously introduced.

\subsection{The sets of runs}

To illustrate the predictability of the learning curves for the
{\sc x}iada corpus, we start with a collection of runs, ${\mathcal C}
=\{{\mathcal E}_i\}_{i \in I}$ generated from the data compiled in
Table~\ref{table-statistics-on-accuracy-no-anchor-5000-700000}. The
latter includes an entry for each one of the learners previously
enumerated, together with its {\sc pl}evel and {\sc cl}evel, as well
as the values for Ac and EAc along the control sequence, from which to
calculate {\sc mape}s, {\sc dmr}s and {\sc rr}s. In order to
improve understanding, all the levels managed are indicated by
their associated word positions in the corpus, which is denoted by
using a superscript {\bf wp} in their identification labels.

\begin{figure}[H]
\begin{center}
\includegraphics[width=0.72\textwidth]{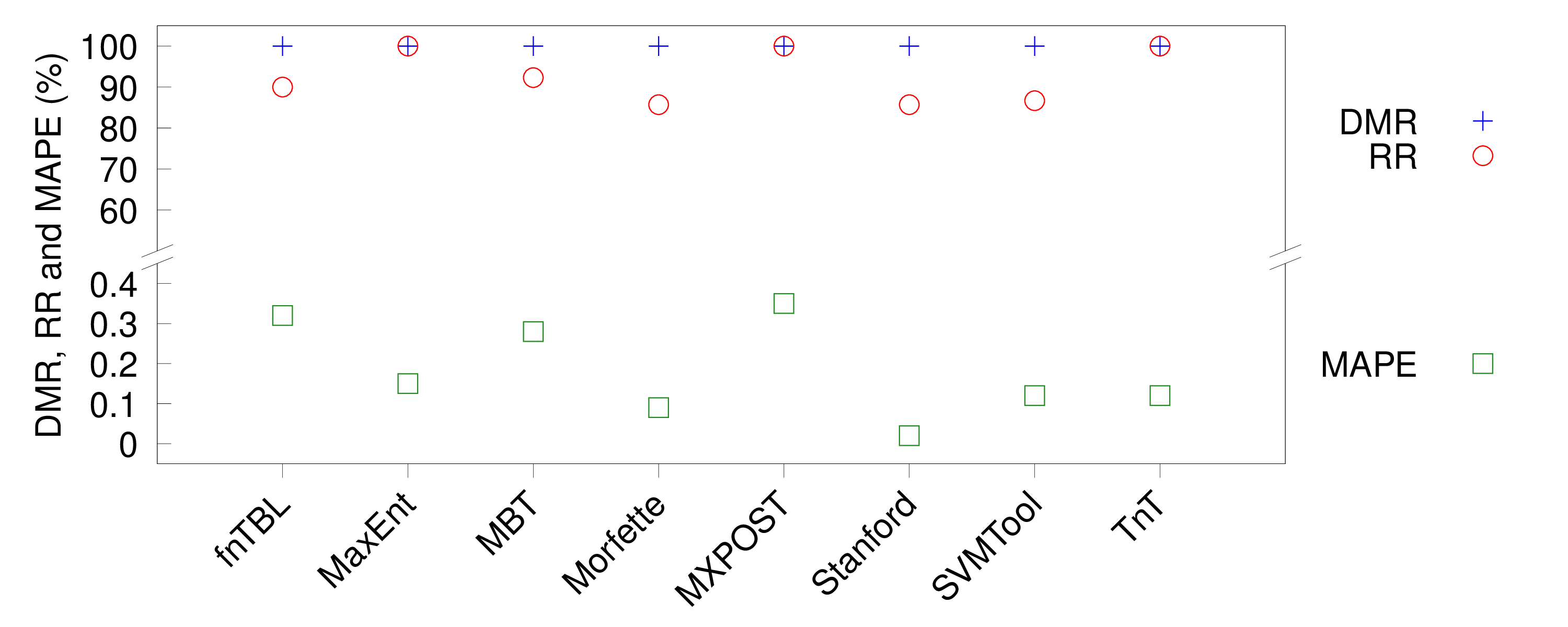} 
\caption{{\sc mape}s, {\sc rr}s and {\sc dmr}s for runs.}
\label{fig-dmr-mape-statistics-no-anchor-5000-700000}
\end{center}
\end{figure}

One detail that attracts our attention is that the run associated
with {\sc t}ree{\sc t}agger does not reach the {\sc pl}evel within
the limits of the training corpus. This behavior is certainly singular
in among all the taggers considered, which highlights the variety
of factors that impact the evaluation of learners, and that, in this
case, leads us to discard considering it in our study. In other
words, in a real model selection process on the {\sc x}iada corpus
selected here, {\sc t}ree{\sc t}agger would not even be placed among
the hypotheses that allow the application of the prediction technique
considered.

\begin{figure}[H]
\begin{center}
  \includegraphics[width=0.7\textwidth]{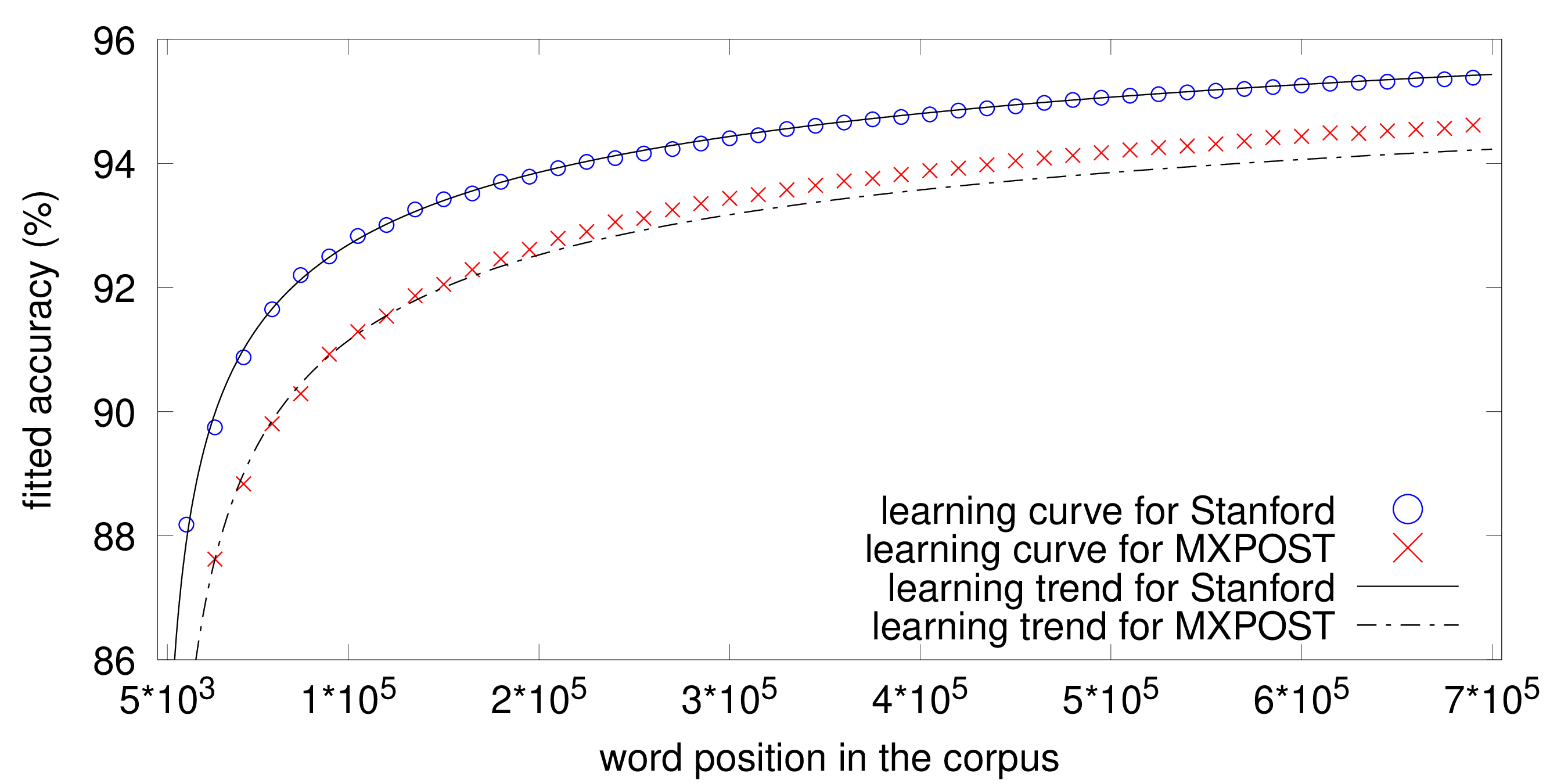}
\caption{Learning trends for the best and worst {\sc mape}s.}
\label{fig-best-worst-MAPE-Ac-EAc-no-anchor-5000-700000}
\end{center}
\end{figure}

\subsection{The quantitative study}

Our reference metric is now the {\sc mape}, whose values are shown
graphically in
Fig.~\ref{fig-dmr-mape-statistics-no-anchor-5000-700000} from the data
compiled in
Table~\ref{table-statistics-on-accuracy-no-anchor-5000-700000}. Taking
into account that we are interested in numbers as small as possible,
the scores range from 0.02 for {\sc s}tanford, to 0.35 for {\sc
  mxpost} in the interval $[\sentencesw{3*10^5},
  \sentencesw{7*10^5}]$. Those results are illustrated in
Fig.~\ref{fig-best-worst-MAPE-Ac-EAc-no-anchor-5000-700000}, showing
the learning curves and learning trends used for prediction on the
runs with best and worst {\sc mape} on the control sequence.  As we
have already done, the observations are generated considering the
portion of the corpus taken from its beginning up to the word position
indicated on the horizontal axis. Finally, 50\% of {\sc mape} values
in this set of runs are in the interval $[0, 0.12]$, a proportion
that reaches 75\% in $[0, 0.28]$. Although these results are slightly
worse than the ones reported in~\cite{VilaresDarribaRibadas16} for
resource-rich languages, they are still very promising, which leads us
to argue for the goodness of the proposal on the quantitative plane.

\subsection{The qualitative study}

Our reference metric is here the {\sc dmr}, whose values are shown
graphically in
Fig.~\ref{fig-dmr-mape-statistics-no-anchor-5000-700000} from the data
compiled in
Table~\ref{table-statistics-on-accuracy-no-anchor-5000-700000}. Taking
into account that we are now interested in scores close to 100, these
range from 71.43 to 100, with 85.71\% of the values in the interval
$[85.71, 100]$. Moreover, the {\sc dmr}s lower than 100 are the result
of the intersection between the {\sc t}n{\sc t} learning curve with
those of {\sc S}tanford and fn{\sc tbl}. Under these conditions, the
maximum value would only be possible if the error in the estimate of
the intersection points was lower than the distance between its
neighbouring control levels, an unrealistic proposition given how
short that distance is (5,000 words). In any case, the results are
comparable to those reported in~\cite{VilaresDarribaRibadas16} for
resource-rich languages, also meeting our expectations from a
qualitative point of view.

\subsection{The study of robustness}

The reference metric is now {\sc rr}, and we are interested in values
as close as possible to 100, the maximum. The results are shown in
Fig.~\ref{fig-dmr-mape-statistics-no-anchor-5000-700000} from the data
compiled in
Table~\ref{table-statistics-on-accuracy-no-anchor-5000-700000}. While
{\sc rr} values range from 85.71 to 100, the latter is only reached in
37.50\% of the runs. This percentage rises to 62.50\% for {\sc rr}s in
the interval $[90,100]$. Overall, these results even exceed those
reported in~\cite{VilaresDarribaRibadas16} for resource-rich
languages, illustrating once again the good performance of the
prediction model, this time against variations in its working
hypotheses.

\section{Conclusions and Future Work}
\label{section-conclusions}

Our proposal arises as a response to the challenge of evaluating {\sc
  pos} tagging models in low-resource scenarios, for which non-deep
learning approaches have often proven to be better suited.
For this purpose, we reuse a formally correct proposal, based on the
early estimation of learning curves. Technically described as the
uniform convergence of a sequence of partial predictors which
iteratively approximates the solution, the method acts as a
proximity condition that halts the training process once a
convergenge/error threshold fixed by the user is reached, and has
already demonstrated its validity when the availability of large
enough learning datasets is not a problem. In order to ensure the
reliability of the results obtained, we have once again used the
testing frame considered then, involving both quantitative and
qualitative aspects, but also the survey of robustness against possible
irregularities in the learning process.

Special attention was paid to the selection of a case study combining
representativeness and access to validation resources, something
somewhat contradictory in the context under consideration. We then
focus on Galician, a minority language of complex morphology, for
which the collection of available training resources is reduced to a
single corpus of sufficient size and quality to ensure both the
validation phase and a rapid convergence process. This set of unique
features allows us to simulate and evaluate short training sessions in
a non-trivial learning environment, associating them with a language
with important deficiencies in terms of computational resources. The
results corroborate the expectations for the theoretical 
basis, placing the performance at a level similar to that observed
in the state-of-the-art for resource-rich languages on the same
learners. This supports the effectiveness of the approach for model
selection considered and its suitability to low-resource scenarios, as
initially argued.

To the best of our knowledge and belief, not only is this the first
time that a proposal for estimating the performance based on the
prediction of learning curves has demonstrated its feasibility in
frameworks of this nature, but it has done so without any type of
prior specific adaptation. In other words, no operational limitations
to the original conceptual design have been observed. All this
justifies the interest in highlighting the independence, both in terms
of language and usage, of the technology deployed. A comprehensible
way of doing it is extending our analysis, first to a broader set of
languages in a variety of language families, and then to other
fundamental and applicative {\sc nlp} tasks, which establishes a clear
line of future work.


\vspace{6pt} 



\authorcontributions{Conceptualization, Manuel Vilares; software,
  V\'{\i}ctor M. Darriba and Francisco J. Ribadas; validation,
  V\'{\i}ctor M. Darriba; investigation, Manuel Vilares and
  V\'{\i}ctor M. Darriba; resources, V\'{\i}ctor M. Darriba, Francisco
  J. Ribadas and Jorge Gra\~na; data curation, V\'{\i}ctor M. Darriba;
  writing---original draft preparation, Manuel Vilares, V\'{\i}ctor
  M. Darriba and Jorge Gra\~na; writing---review and editing, Manuel
  Vilares and V\'{\i}ctor M. Darriba; visualization, Manuel Vilares
  and V\'{\i}ctor M. Darriba; supervision, Manuel Vilares; project
  administration, Manuel Vilares; funding acquisition, Manuel
  Vilares. All authors have read and agreed to the published version
  of the manuscript.}

\funding{This research was partially funded by the Spanish Ministry of
  Science and Innovation through projects PID2020-113230RB-C21
  and PID2020-113230RB-C22, and by the Galician Regional Government
  under project ED431C 2020/11.}



\institutionalreview{Not applicable.}

\informedconsent{Not applicable.}
  
\dataavailability{Not applicable.}
  
\conflictsofinterest{The authors declare no conflict of interest.}



\abbreviations{Abbreviations}{
  The following abbreviations are used in this manuscript:\\
  
\noindent 
\begin{tabular}{@{}ll}
{\sc a}c & Accuracy \\
{\sc al} & Active Learning \\
{\sc cl}evel & Convergence Level \\
{\sc dl} & Deep Learning \\
{\sc dmr} & Decision-Making Reliability \\
{\sc ea}c & Estimated Accuracy \\
{\sc hmm} & Hidden M\'arkov Model \\
{\sc mem} & Maximum Entropy Model \\
{\sc mbt} & Memory-Based Tagger \\
{\sc ml} & Machine Learning \\
{\sc nlp} & Natural Language Processing \\
{\sc pe} & Percentage Error \\
{\sc pos} & Part-of-Speech \\
{\sc pl}evel & Prediction Level \\
{\sc re} & Reliability Estimation \\
{\sc rer} & Reliability Estimation Ratio \\
{\sc rr} & Robustness Rate \\
{\sc svm} & Support Vector Machine \\
{\sc wl}evel & Working Level
\end{tabular}
}

\begin{adjustwidth}{-\extralength}{0cm}

\reftitle{References}

\end{adjustwidth}
\end{document}